# A New Method for Cross-Lingual-based Semantic Role Labeling

Mohammad Ebrahimi[1], Behrouz Minaei Bidgoli[1] and Nasim Khozouei[2]

[1]School of Computer Engineering, Iran University of Science and Technology, Iran
[2]Department of Computer Engineering, Faculty of Engineering, Yasuj University, Iran

## Abstract

Semantic role labeling is a crucial task in natural language processing, enabling better comprehension of natural language. However, the lack of annotated data in multiple languages has posed a challenge for researchers. To address this, a deep learning algorithm based on model transfer has been proposed. The algorithm utilizes a dataset consisting of the English portion of CoNLL2009 and a corpus of semantic roles in Persian. To optimize the efficiency of training, only ten percent of the educational data from each language is used. The results of the proposed model demonstrate significant improvements compared to Niksirt et al.'s model. In monolingual mode, the proposed model achieved a 2.05% improvement on F1-score, while in cross-lingual mode, the improvement was even more substantial, reaching 6.23%. Worth noting is that the compared model only trained two of the four stages of semantic role labeling and employed golden data for the remaining two stages. This suggests that the actual superiority of the proposed model surpasses the reported numbers by a significant margin. The development of cross-lingual methods for semantic role labeling holds promise, particularly in addressing the scarcity of annotated data for various languages. These advancements pave the way for further research in understanding and processing natural language across different linguistic contexts.

## Introduction

Semantic role labeling (SRL) is an essential part of natural language processing that aims to assign specific roles to different parts of a sentence, enabling a deeper understanding of language. Semantic role labeling is often described as the process of automatically answering questions such as "who did what, to whom, when, where, why, and how.". SRL has multiple applications in natural language processing, including multi-document summarization [1], question answering [2], reading comprehension [3], sentiment analysis [4], and caption generation [5]. Initially, researchers tackled this problem by manually creating various rules and applying them to sentences in a specific language. However, the introduction of traditional machine learning techniques, such as support vector machines and simple Bayes, significantly improved the results in this domain. This approach to solving SRL involved extracting information like syntactic information manually. Researchers then utilized various machine learning algorithms, such as decision trees, support vector machines, and Bayes, to extract predicates, and arguments and categorize semantic roles.

However, while traditional machine learning methods have shown promise, recent advancements in deep learning and neural networks have provided even more accurate and efficient solutions to SRL. These approaches take advantage of large annotated datasets and leverage the power of neural architectures, such as recurrent neural networks (RNNs), Pretrained language models, attention and self-attention mechanisms, graph neural networks, and sequenceto- sequence models to capture contextual dependencies and improve semantic role labeling performance. The combination of these advancements has enhanced the accuracy and effectiveness of NLP systems, leading to exciting possibilities for further research and applications. With the continuous development of SRL algorithms, we can expect further advancements in natural language understanding, enabling more sophisticated applications in various fields, including information retrieval, machine translation, and dialog systems.

Many semantic role labeling models have been developed with a focus on specific languages, making them monolingual. These models are designed with assumptions and dependencies on the language of the input sentences. Another approach is to develop multilingual semantic role labeling systems. The goal of multilingual SRL is to identify and classify the various semantic roles in a sentence or phrase, regardless of the language being used. To achieve this, scientists have explored different approaches. Some have adapted existing models to handle multiple languages by incorporating language-specific features or developing language-specific SRL systems. Others have developed more language-agnostic models that can handle multiple languages without any finetuning or language-specific modifications. The lack of high-quality data resources in many languages has been a significant challenge in semantic role labeling. To address this issue, researchers in natural language processing have started exploring crosslingual semantic role labeling methods. Here are some approaches in this field:

- Annotation projection-based approaches: This method relies on a large parallel text corpus between the source language and the target language. sentences in the source language are initially annotated with semantic role labels using a source Semantic Role Labeling system. To extend these annotations to the sentences in the target language, word alignment techniques are employed. By aligning the words in parallel sentences, the labels from the origin language are projected onto the equivalent words in the second language sentences [6] [7].
- Translation-based approaches: Recent advancements in neural machine translation [8] have opened up new possibilities for cross-lingual transfer. Translation-based approaches aim to to minimize noise by translating gold data directly into the desired language. This approach leverages the power of neural machine translation to improve semantic role labeling across languages [9] [10].
- Model transfer approaches: In this approach, a source language model is modified to be applied directly to a new language. Multilingual models are built using language-independent features like cross-language word representations and universal part-of-speech labels, which can be transferred to target languages. This allows for the transfer of knowledge from a source language to a new language for semantic role labeling [11] [12] [13] [14].

## Related Work

### Syntactic Information

Semantic role labeling models can be categorized as either syntax-aware or syntax-agnostic, based on their utilization of syntactic information. Traditional approaches relied on manually extracted features, such as syntactic information, as sentence features. However, with the development of deep learning models, researchers started relying on automatic feature extraction and initially abandoned the use of syntactic information [15] [16] [17]. Nevertheless, some researchers believe that incorporating syntactic information into the input of deep learning models can enhance their performance. Consequently, using or not using syntactic information has become a research challenge discussed in numerous papers [18] [19] [20].

**Semantic Role Labeling Styles**

Semantic role labeling is done in two main styles:

- Framenet style: presented in 1998 by Baker et al [21].
- Propbank style presented in 2005 by Palmer et al [22].

In the Propbank style, predicates and their arguments are represented using two approaches: spanbased and dependency-based.

- Span-based semantic role labeling: first presented at the Computer Natural Language Learning conference in 2004 [23] and 2005 [24] , utilizes a BIO representation. This means that the labeling is done by specifying the beginning (B) and inside (I) boundaries of the argument spans within the sentence.
- Dependency-based semantic role labeling: introduced at the 2008 [25] and 2009 [26] Computational Natural Language Learning conference, captures the relationships between words in a sentence using dependency structures. This style focuses on identifying the roles that words play in relation to the main predicate.

**End-to-End Semantic Role Labeling**

The pipeline of semantic role labeling consists of four main stages: predicate identification, predicate sense disambiguation, argument identification, and argument classification. Initial research efforts focused on addressing each stage separately [27] [28] [29]. Recent studies have successfully demonstrated the benefits of tackling some of these sub-tasks jointly through multitask learning [30]. This means that when all four stages are performed by the model together, the algorithm is referred to as end-to-end.

The advantage of an end-to-end SRL approach is that it enables the model to capture the dependencies and interactions between different subtasks, leading to better overall performance. By jointly learning these tasks, the model can leverage the shared knowledge and patterns present in the data, enhancing its ability to understand the underlying semantic relationships between predicates and their arguments.

However, it is important to note that implementing an end-to-end SRL system can pose challenges, such as dealing with the increased complexity and potential error propagation across the various subtasks. Nonetheless, current research efforts in multitasking learning have demonstrated promising results, highlighting the potential benefits of integrating these subtasks into a unified framework.

**Monolingual Semantic Role Labeling**

Monolingual semantic role labeling refers to models that are trained and applied in a single language, such as English. These models are primarily designed to understand the semantic roles of words and phrases within a sentence in that particular language. While monolingual SRL models are effective for the language they are trained on, they may not generalize well to other languages due to differences in grammar, syntax, and semantics.

**Multilingual Semantic Role Labeling**

Multilingual semantic role labeling (SRL) has gained significant attention since the introduction of the CoNLL-2009 multilingual annotated dataset in 2009. The dataset included six languages: Catalan, Spanish, English, Czech, German, and Chinese, and served as a foundation for research in this field. The primary objective of multilingual SRL is to develop language-independent models capable of generating reliable results across different languages. In 2019, He, Li, and Zhao [31] proposed a multi-language model that relied on synthetic information. This approach aimed to bridge the language gap by generating artificial data to improve model performance across multiple languages. However, a year later, Conia and Navigili [11] highlighted a potential limitation of this approach. They argued that syntactic information varies significantly across different languages in terms of morphology and structure, creating obstacles to the effectiveness of multilingual models. To overcome this limitation, Conia and Navigili introduced their multilingual model, which did not rely on syntactic information. By eliminating the use of syntactic features, this model aimed to achieve language independence more effectively. Their approach likely focused on capturing semantic dependencies and role assignments without explicitly considering the specific syntactic characteristics of each language. Overall, the field of multilingual SRL continues to evolve as researchers explore different approaches and techniques to create language-independent models capable of performing well across diverse languages.

**Cross-lingual Semantic Role Labeling**

Cross-lingual semantic role labeling (SRL) is a field that aims to overcome the challenge of insufficient training data for building SRL systems. Since semantic role annotations are currently only accessible for a small number of languages, researchers have been focusing on leveraging resources from a source language to reduce the effort and expense associated with creating models or annotations for a new target language. Recent advancements in representation learning have aided this goal. By employing techniques that enable models to understand the underlying semantics of text, researchers can use these models trained on source language data to perform SRL in a target language. This approach helps in overcoming the scarcity of annotated data for the target language. Since most of the research conducted in recent years has been dedicated to Cross-lingual semantic role labeling, we will introduce some of the suggested methods:

**Annotation Projection:** implementing crosslingual semantic role labeling faces the challenge of dissimilar predicate senses and semantic role sets across languages. This issue is evident in datasets like CoNLL-2009, which consists of multiple datasets for various languages such as English, Chinese, Catalan, Spanish, German, and Czech. These datasets often lack alignment as they are developed based on different linguistic theories. Consequently, this misalignment introduces additional challenges for cross-lingual semantic role labeling systems. Annotation projection is a common approach used in natural language processing. It relies on a large parallel text corpus that includes both the source and target languages. In this approach, the source language sentences are first automatically annotated with semantic role labels using a source semantic role labeler. These annotations are then projected onto the target side sentences based on word alignment between the source and target languages. This technique helps in leveraging the existing annotations to generate semantic role labels for the target language without requiring manual annotation [6] [7]. The work by Pado and Lapata [32], and Akbik et al. [33] focused on addressing these issues. They proposed the English PropBank, which consists of a set of predicates and a universal semantic role. This provided a framework for annotating semantic roles in English sentences.To extend this annotation to non- English sentences, they employed word alignment techniques commonly used in parallel collections, like Europarl [34]. These efforts led to the development of the Universal Propbank [35], which is a multilingual compilation of semi-automatic annotation sets designed for semantic role labeling. It is actively used today for the purpose of training and assessing cross-lingual methods, including word alignment techniques [6]. The quality of parallel data can indeed impact the performance. If the data contains inaccuracies or inconsistencies, it may introduce noise and affect the accuracy of the model. Similarly, the performance of the source language semantic role labeling model plays a crucial role. If it produces incorrect or incomplete annotations, it can hinder the quality of the generated translations. Additionally, the accuracy of alignment tools used to align source and target sentences is important. If the alignment is not precise, it may introduce errors or misalignments, leading to noise. Building a semantic role labeling model in Farsi using this method does require parallel corpora of Persian and English. However, constructing such corpora can indeed be challenging and costly.

**Translation-based Models:** Translation-based approaches to semantic role labeling have gained significant traction due to the advancements in neural machine translation [8]. These approaches offer the potential for high cross-lingual transfer capabilities. The primary objective of translation-based approaches is to mitigate the noise caused by the original labeler by translating the high-quality data directly into the desired language [9] [10]. In cases where parallel corpora are not available, annotation scheme techniques can still be utilized by automatically translating an annotated corpus and subsequently mapping the original labels onto the translated corpus. Daza and Frank have recently succeeded in training an sequence to sequence architecture to jointly deal with semantic role labeling and translation [9].

Both of the previously proposed approaches have certain limitations. The assumption that the word arrangement is the same in the source and target languages is a crucial defect. Languages vary in their word order and sentence structure, so a direct word-for-word mapping may not be accurate. Additionally, relying on annotations based on English can be limiting. Different languages have their unique linguistic features and nuances that may not be properly captured when using English annotations. It is important to consider the specific characteristics and requirements of each language when developing translation methods.

**Model transfer:** Model transfer is a powerful approach in natural language processing that enables the application of a source language model to a new language [11] [12] [13] [14]. This involves modifying a source language model so that it can be directly applied to the target language. One way to achieve model transfer is by building multilingual models that rely on language-independent features, such as cross-lingual word representations and universal part-of-speech labels. These features can be transferred directly to target languages, allowing the model to work effectively across different languages. Multi-task learning can also be utilized within this approach, where the model is simultaneously trained on several languages. In the case of semantic role labeling, multi-task learning can be applied to parallel, translated, and aligned corpora. For example, the universal Propbank corpus can be utilized for this purpose. However, Conia and Navigili [14] have proposed a model that doesn't rely on parallelism or the uniformity of semantic role labels in different languages. Instead, their model employs universal encoders to learn a unified representation that can be shared across multiple languages. Overall, model transfer is a promising approach for building multilingual models that can be directly applied to new languages, leveraging language-independent features and multi-task learning techniques.

The proposed model belongs to the last approach type, known as model transfer. It utilizes two universal sentence encoders and universal predicateargument encoders that are language-independent. Multi-task learning is employed in training this model, allowing for simultaneous updates of the common and non-common parts across different languages. This is end-to-end, meaning that all stages of the semantic role labeling pipeline are learned together without relying on language-specific syntactic information. Finally, it is necessary to mention that we use the algorithm proposed by Conia et al. [14] to perform cross-lingual semantic role labeling. Here is the link to the code https://github.com/SapienzaNLP/unify-srl.

**Model**

This model is capable of simultaneously handling predicate identification, predicate sense disambiguation, argument identification, and argument classification, so it falls under the extensive group of end-to-end systems; The model's architecture can be roughly partitioned into the following elements:

- A universal sentence encoder that shares its parameters among languages. It generates word encodings that capture information related to predicates.
- A universal predicate-argument encoder that also shares its parameters among languages. It models the relationships between predicates and arguments.
- A set of language-specific decoders that perform the following tasks: determining whether words are predicates or not, selecting the most appropriate sense for each predicate, and assigning a semantic role to each predicate-argument pair. These decoders consider various semantic role labeling inventories.

This model stands out from previous studies by eliminating the need for cross-source mapping, word alignment techniques, translation tools, other annotation transfer techniques, or parallel data in order to achieve accurate semantic role labeling. It achieves this solely through cross-linguistic implicit knowledge transfer. Figure 1 shows the architecture of the proposed model.

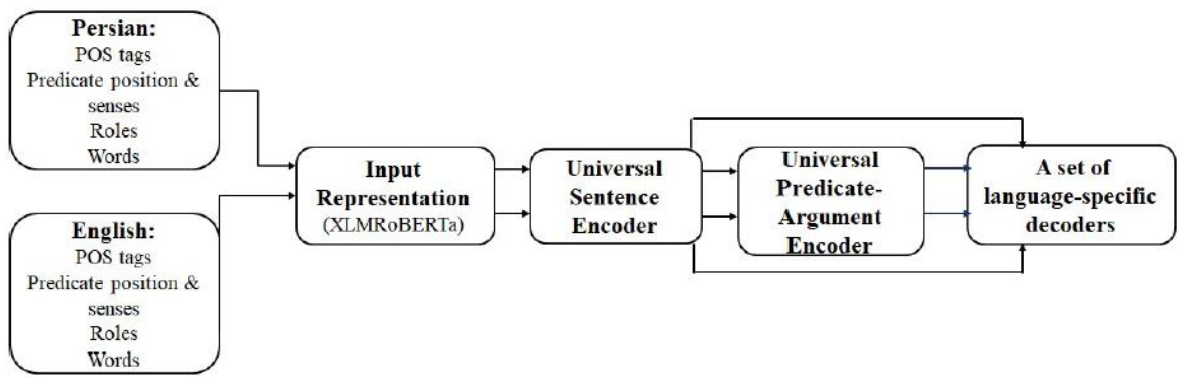

**Figure 1:** The architecture of the proposed model

**Input Representation**

The input representation in pre-trained linguistic models like ELMo, BERT, and XLMRoBERTa has become the standard due to their capability to encode extensive knowledge. Recent research, such as Hewitt and Manning [36] and Conia and Navigli [13], has shown that distinct levels of a language model encompass diverse syntactic and semantic characteristics. In this model, we join the hidden states from the top four inner layers of the language model to construct a textual representation for each input word.

In Figure 2, we illustrate the architecture of a 12- layer BERT model. For us, the word embedding is obtained by concatenating the hidden states from layers 9, 10, 11, and 12, as shown in Figure 3.

$w = (w_0, w_1, \ldots, w_i, \ldots, w_{n-1})$, a specific sentence

$h_i^k = l^k(w_i|w)$, hidden state, where $l^k$ represents the internal computation in the language model with K layers

We compute the encoding model $e_i$ as follows:

$$h_i = h_i^k \oplus h_i^{k-1} \oplus h_i^{k-2} \oplus h_i^{k-3}$$
$$e_i = \text{Swish}(W^w h_i + b^w)$$

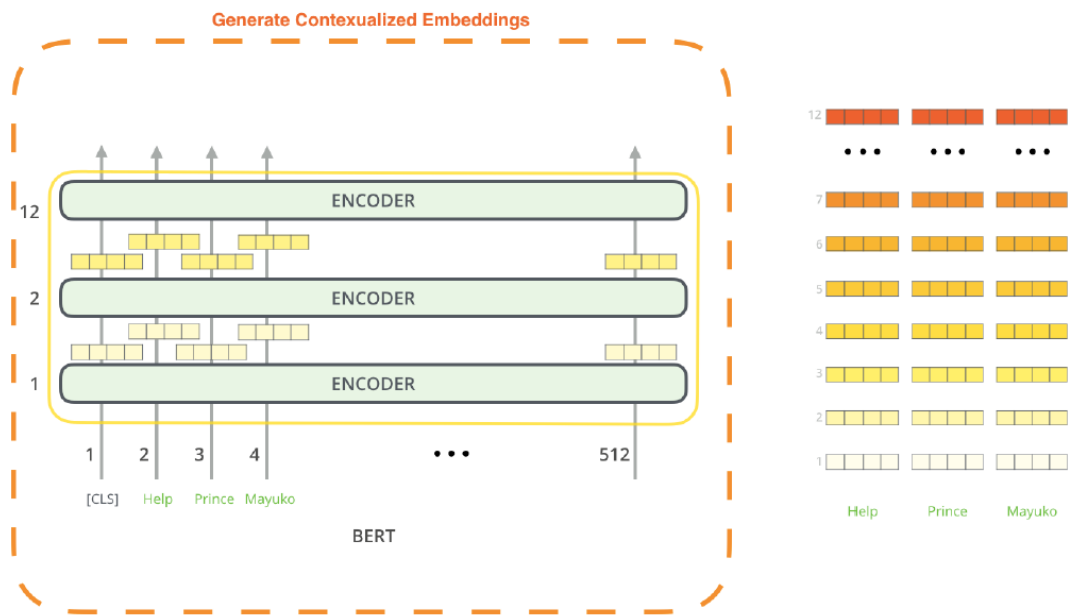

Figure 2: BERT architecture with 12 encoders

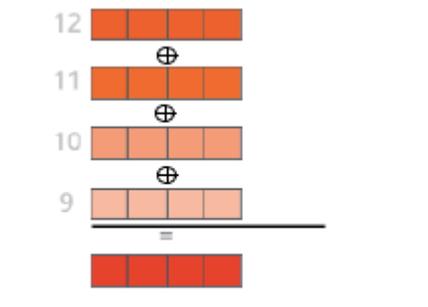

Figure 3: Concatenation of the last four layers of BERT

x ⊕ y = Concatenation of two vectors x and y.

The selection of an activation function is indeed crucial for the dynamics of training and performance in deep neural networks. While ReLU (f(x) = max(0,x)) has been widely successful, alternative activation functions have struggled to consistently outperform it. However, the Google Brain team has proposed a new activation function called Swish (f(x) = x · sigmoid(x)).

Swish offers a smoother slope compared to the older ReLU function [37]. This smoothness helps in addressing some of the issues associated with ReLU, such as the "dying ReLU" problem. Swish function's non-uniformity allows it to handle a wider range of data and adapt better to different types of inputs. The diagram in Figure 4 illustrates the Swish activation function visually. The Google Brain team believes that Swish could be a promising alternative to ReLU and may provide additional benefits in certain scenarios.

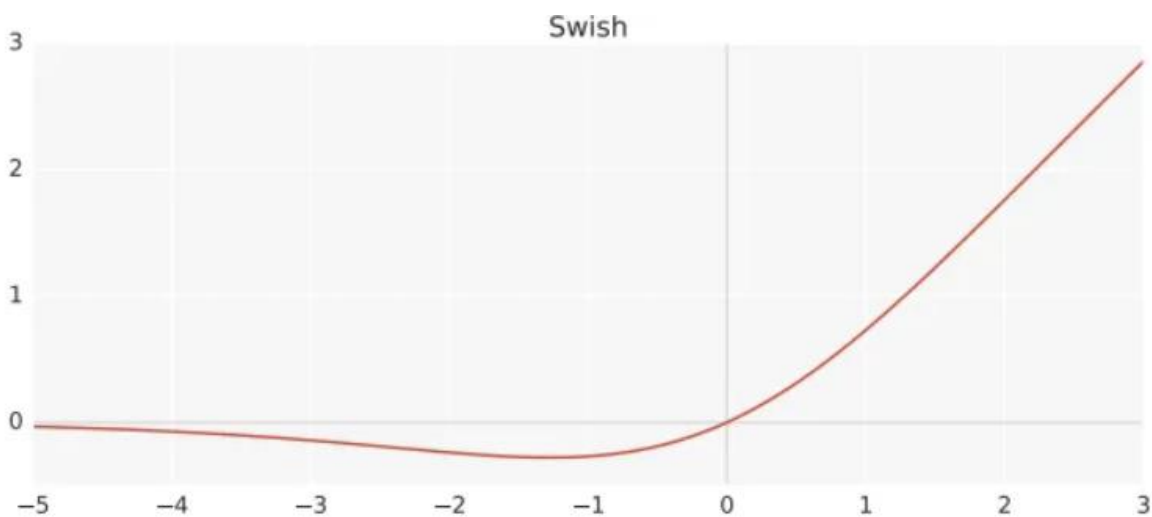

Figure 4: Swish activation function

**Multilingual BERT:** Multilingual BERT is a pretrained language model that encompasses 104 different languages. It has been trained extensively on a wide range of multilingual data derived from Wikipedia The training process follows a selfsupervised approach, wherein the model is fed with raw texts and generates inputs and labels automatically, without the need for human annotations. This allows the model to utilize publicly available data effectively.

BERT is trained on two primary tasks: The first is Masked Linguistic Modeling (MLM), where a sentence is randomly masked by hiding 15% of its words. The model then attempts to predict the hidden words by processing the entire masked sentence. This differs from traditional models like Recurrent Neural Networks (RNNs) or GPT, which usually see words in sequence or hide tokens in subsequent steps. MLM enables BERT to learn a bidirectional representation of the sentence. The second task is Next Sentence Prediction (NSP), where pairs of masked sentences are provided as input during pre-training. These sentence pairs may or may not be adjacent in the original text. The objective is for the model to predict whether the two sentences were originally sequential or not. By undergoing these two training tasks, BERT gains a comprehensive understanding of the linguistic structure and context in various languages, making it highly effective for multilingual natural language processing tasks.

**XLM-RoBERTa**: XLM-RoBERTa [38] is a variant of RoBERTa that has been pre-trained on a massive amount of multilingual data. It has received specialized training using 2.5 terabytes of meticulously processed CommonCrawl data, which includes text from 100 different languages. RoBERTa itself is a transformer-based model that has been pre-trained in a self-supervised manner. When we say "self-supervised," it means that the model has been trained only on raw texts without any human labels. Instead, an automatic process is used to generate inputs and labels from these texts. This approach allows the model to utilize publicly available data and doesn't require human annotation. RoBERTa underwent training through the masked language modeling task, wherein specific words within the input text were masked, and the model made predictions for these masked words. This process helps the model learn an internal representation of 100 languages, enabling it to extract useful features for downstream tasks. For example, if you have a dataset of labeled sentences, you can use the features generated by XLM-RoBERTa as input to a standard classifier during fine-tuning. While XLMRoBERTa can be used for masked language modeling, its primary purpose is to be fine-tuned for specific downstream tasks. Among the two pre-trained models mentioned above, we opted to use XLM-RoBERTa. This choice was made based on its superior accuracy in various tasks compared to Multilingual BERT. The key differentiating factor is that XLM-RoBERTa utilizes RoBERTa instead of BERT. To validate our claim, we conducted practical evaluations and found that while these two models exhibit minimal differences with large training datasets, the disparities become more apparent, particularly in cross-lingual scenarios involving the Persian language. As a result, we proceeded with conducting our experiments exclusively using XLM-RoBERTa.

**Universal Sentence Encoder**
Building upon Fillmore's original intuition [39] regarding the presence of profound semantic relationships between predicates and other constituents within a sentence, Conia proposes the notion that these semantic relations can be preserved across different languages. Guided by this notion, he introduce a universal sentence encoder that utilizes shared parameters between languages. The main goal of this universal sentence encoder is to capture language-independent sentence-level information, such as predicate position and sense, that remains consistent across various languages. This implementation of the universal sentence encoder follows a similar approach to Marchegiani et al. [16], Cai et al. [40], and He et al. [31], employing a stack of Bidirectional LSTM layers. However, he address the issue of gradient vanishing by connecting the output of each layer to its input. In other words, given a sequence of word encodings e = ($e_0$, $e_1$, . . . , $e_i$, . . . , $e_{n-1}$), the model calculates a sequence of encodings of time step t as follows:

$$t_i^j = \begin{cases} \mathbf{e}_i & \text{if } j = 0 \\ \mathbf{t}_i^{j-1} \oplus \text{BiLSTM}_i^j(\mathbf{t}^{j-1}) & \text{Otherwise} \end{cases}$$

$$\mathbf{t} = \langle \mathbf{t}_0^{K'}, \mathbf{t}_1^{K'}, \ldots, \mathbf{t}_{n-1}^{K'} \rangle$$

The i-th time step of the j-th BiLSTM layer is denoted as $\text{BiLSTM}_i^j(.)$, while K' represents the total number of stacked layers. At each encoding time step $t_i$, the model initiates by generating a predicate representation $p_i$ to indicate whether the corresponding word $w_i$ is a predicate. Furthermore, it produces a sense representation $s_i$ that captures information about the sense of a predicate at position i.

$$\mathbf{p}_i = \text{Swish}(\mathbf{W}^p \mathbf{t}_i + \mathbf{b}^p)$$
$$\mathbf{s}_i = \text{Swish}(\mathbf{W}^s \mathbf{t}_i + \mathbf{b}^s)$$

The vector representations obtained for each time step, predicate, and predicate sense are situated within three shared spaces across different languages and formalisms employed for semantic role labeling implementation.

## Universal Predicate-Argument Encoder

In accordance with the design of the universal sentence encoder, this model integrates a universal predicate-argument encoder that shares parameters across different languages. The purpose of this supplementary encoder is to capture the connections between predicate-argument pairs in a language-neutral manner. Like the universal sentence encoder, the universal predicate-argument encoder is constructed using a stack of BiLSTM layers. In more exact terms, when $w_p$ symbolizes a predicate within the input sentence $w = (w_0, w_1, \ldots, w_p, \ldots, w_{n-1})$, the model produces a series of encodings for the particular argument of predicate a through the following process:

$$\mathbf{a}_i^j = \begin{cases} \mathbf{t}_p \oplus \mathbf{t}_i & \text{If } j=0 \\ \mathbf{a}_i^{j-1} \oplus \text{BiLSTM}_i^j(\mathbf{a}^{j-1}) & \text{Otherwise} \end{cases}$$

$$\mathbf{a} = \langle \mathbf{a}_0^{K''}, \mathbf{a}_1^{K''}, \ldots, \mathbf{a}_{n-1}^{K''} \rangle$$

Within this context, the model produces a semantic role representation $r_i$ for the word $w_i$ by leveraging the i-th encoded time step $t_i$ from the universal sentence encoder and the overall number of stack layers K". The initiation of this process involves the encoding of each sentence-specific argument $a_i$.

$$\mathbf{r}_i = \text{Swish}(\mathbf{W}^r \mathbf{a}_i + \mathbf{b}^r)$$

Just like the predicate and sense representations p and s, the semantic role r should make use of cross-linguistic information to abstract language-specific features, given that the predicate-argument encoder is consistent across all languages..

## Language-Specific Decoders

The encodings for predicates, senses, and semantic roles mentioned earlier are consistent across languages, requiring the model to focus on semantic features rather than surface-level ones like word order, part-of-speech tags, and syntactic rules that can vary between languages. However, the ultimate goal is for our model to offer semantic role annotations based on established predicate-argument structure databases like PropBank, Ancora, or PDTVallex. As a result, the proposed model comprises a series of linear decoders that determine whether a word $w_i$ is a predicate, the most suitable sense of the predicate $w_p$, and the semantic role of the word $w_r$ in relation to a specific predicate $w_p$. For each language l:

$$\sigma^P(w_i|l) = \mathbf{W}^{p|l}\mathbf{p}_i + \mathbf{b}^{p|l}$$
$$\sigma^s(w_p|l) = \mathbf{W}^{s|l}\mathbf{s}_i + \mathbf{b}^{s|l}$$
$$\sigma^r(w_r|w_p, l) = \mathbf{W}^{r|l}\mathbf{r}_i + \mathbf{b}^{r|l}$$

While more intricate decoding strategies were an option, linear decoders offer two benefits in our scenario: firstly, they maintain the simplicity of the language model component and promote learning from the model's universal encoders. Secondly, they act as linear explorers, offering valuable insights into the model's ability to acquire cross-lingual knowledge.

**Training objective**

The objective of this training is to minimize the overall batch classification errors by training the model in a multitasking manner. This includes reducing cross-entropy in predicate identification, predicate sense disambiguation, and argument identification/classification across all languages. In a more formal manner, taking into account language l and the associated predicate identification error $\ell^{p|l}$, predicate sense disambiguation error $\ell^{s|l}$, and argument identification/classification error $\ell^{r|l}$, the total error $\ell$ can be expressed as:

$$\mathcal{L} = \sum_{l \in L} \left( \mathcal{L}^{p|l} + \mathcal{L}^{s|l} + \mathcal{L}^{r|l} \right)$$

where L is the set of languages—and corresponding formalisms—in the training set.

**Fine Tuning**

Fine-tuning involves the precise application of transfer learning techniques. Essentially, it entails adjusting a pre-trained model to suit a related task, thereby optimizing its performance. Assuming that the primary task is similar to the new task, using a predesigned and pre-trained artificial neural network allows us to leverage what the model has previously learned without the need to develop it from scratch.

BERT, developed by Google, was a paradigm shift in natural language modeling, particularly due to the introduction of pre-training/fine-tuning. After pre-training in an unsupervised manner on a massive amount of textual data, the model can quickly be finetuned on a specific downstream task with relatively few labels, as it has already learned general language patterns during the training process. The intuition behind BERT is that the initial layers learn general language patterns that have some relevance to the downstream task, while the subsequent layers learn task-specific patterns. This intuition aligns with deep computer vision models, where the initial layers learn general features like edges and corners, and the later layers learn specific features like eyes and noses for face detection.

One of the fine-tuning techniques, known as partial freezing, involves keeping the initial layers of BERT fixed, meaning they are not updated during the fine-tuning process, while modifying the higher layers. In our research, we perform both fine-tuning with and without partial freezing. For cross-lingual semantic role labeling with fine-tuning, we follow the approach mentioned. This means we keep the initial layers of xlm-roberta-base fixed and allow the parameters of the top 4 layers to be updated during the training process.

**Datasets**

Our proposed model is simultaneously trained on two parts of the CoNLL-2009 English dataset and the Persian PropBank corpus. In the original research, the model was trained on the complete CoNLL-2009 dataset using data from multiple languages, resulting in significant performance. However, due to the unavailability of the complete dataset, we only use the English portion of the CoNLL-2009 dataset. In the Persian dataset, the samples are allocated as follows: 10% for training data, 10% for validation data, and the remaining samples for testing data. The designed algorithm is trained and tested on these two datasets, and the results are reported using the F1 metric. The statistics of the CoNLL-2009 dataset and the Persian PropBank corpus are shown in Table 1 and Table 2, respectively.

Table 1: Statistical characteristics of CoNLL-2009 dataset

| Language | Training Samples | Validation Samples | Test Samples |
|---|---|---|---|
| Catalan | 13200 | 1724 | 1862 |
| Czech | 38727 | 5228 | 4213 |
| German | 36020 | 2000 | 2000 |
| English | 39279 | 1334 | 2000 |
| Spanish | 14329 | 1655 | 1725 |
| Chinese | 22277 | 1762 | 2556 |

Table 2: Statistical characteristics of the Persian PropBank dataset

| Training Samples | Validation Samples | Test Samples |
|---|---|---|
| 23984 | 2999 | 2998 |

**Results**

The proposed method is the first cross-lingual learning algorithm applied to the Persian language. Since the algorithm is trained on multiple languages and in a multitask manner, we can only utilize pretrained multilingual models such as Multilingual BERT and XLM-RoBERTa. However, XLM-RoBERTa model outperforms significantly on cross-lingual tasks. Additionally, the available hardware resources in terms of memory and GPU only allowed us to execute this model. Therefore, we have only used XLM-RoBERTabase. By using only 10% of the Persian training dataset and the complete English dataset, we were able to improve the F1 score from 71.76 (in monolingual setting) to 75.94 (in cross-lingual setting), resulting in a 4.18% improvement.

Since no research has been conducted on crosslingual semantic role labeling in Persian, we ran the best Persian semantic role labeling model [38] on 10% of the training data. The F1 score for this model was 69.71. Finally, our proposed model performs 2.05% better in the monolingual setting and 6.23% better in the cross-lingual setting compared to this model. However, we believe that the difference between these two models is greater than 6.23%, as our proposed model performs all four stages of the semantic role labeling pipeline jointly, while the other model excludes the first two stages and assumes that they have been performed by another machine.

**Experiments**

To evaluate the proposed semantic role labeling model in resource-constrained settings, similar to our original paper, we use 10% of the training data for each language, namely Persian and English. To better compare and demonstrate the performance of our proposed model, we evaluate it in various settings, including monolingual and cross-lingual settings.

In Table 3, the F1 score results of the proposed model are reported on ten percent of the Persian language dataset, with varying amounts of English data from zero to one hundred percent, in both monolingual and cross-lingual settings. Please note that the monolingual setting refers to the model being trained and evaluated solely on one language, which means that in our experiments, the English data volume is zero percent. The cross-lingual setting occurs when two or more languages undergo multi-task learning during the training phase, and eventually, the model is evaluated on those languages. This refers to the next ten experiments in which we gradually increase the English data volume from ten percent to one hundred percent.

Table 3: *Results of cross-lingual semantic role labeling on 10% of the Persian data*

| English percentage | F1 | Precision | Recall |
|---|---|---|---|
| 0 | 71.76 | 72.53 | 70.99 |
| 10 | 74.11 | 74.59 | 73.64 |
| 20 | 75.1 | 75.44 | 74.76 |
| 30 | 74.95 | 75.67 | 74.24 |
| 40 | 75.43 | 76.09 | 74.78 |
| 50 | 75.51 | 75.94 | 75.08 |
| 60 | 75.79 | 76.22 | 75.34 |
| 70 | 75.43 | 76.29 | 74.59 |
| 80 | 75.78 | 76.24 | 75.32 |
| 90 | 75.86 | 76.32 | 75.4 |
| 100 | 75.94 | 76.41 | 75.47 |

The highest F1 for Persian in cross-lingual mode is 75.94.

The proposed model is the first model in the field of cross-lingual semantic role labeling that is implemented for the Persian language. Therefore, there is no possibility to compare its results with any previous research in the field of cross-lingual semantic role labeling. As a result, we compare our proposed model's performance with the study by Nikseirat et al. [41], which has the best results in semantic role labeling for Persian on all data. Please note that our proposed model encompasses all four stages of the end-to-end semantic role labeling pipeline, while the model by Nikseirat et al. [41] only considers the first two stages, which are predicate identification, predicate sense disambiguation provided from gold data. These types of models assume that the first two stages of the pipeline have already been performed by other models and the output is fed into this model.

Table 4: *The results of semantic role labeling by Nikseirat et al. on 10% of the data.*

| F1 | Precision | Recall |
|---|---|---|
| 69.71 | 78.53 | 62.67 |

F1 score of the research by Nikseirat et al. [38] is 69.71.

Based on the observations, incorporating 10% of the English dataset with 10% of the Persian dataset results in a 2.35% improvement in the F1 measure. While we expected a more significant impact from increasing the amount of English data, this improvement is not substantial. However, it should be noted that the improvement increases to 4.18%, indicating a 1.83% enhancement compared to the original paper's results. In terms of comparing the proposed method with the study by Nikseirat et al., the proposed model outperforms the compared model by 2.05% in the monolingual mode and 6.23% in the multilingual mode. It is worth mentioning that the compared model only learns the semantic labeling role in two out of the four stages and relies on golden data for the other two stages. Therefore, the actual superiority of the proposed model is significantly higher than the mentioned numbers.

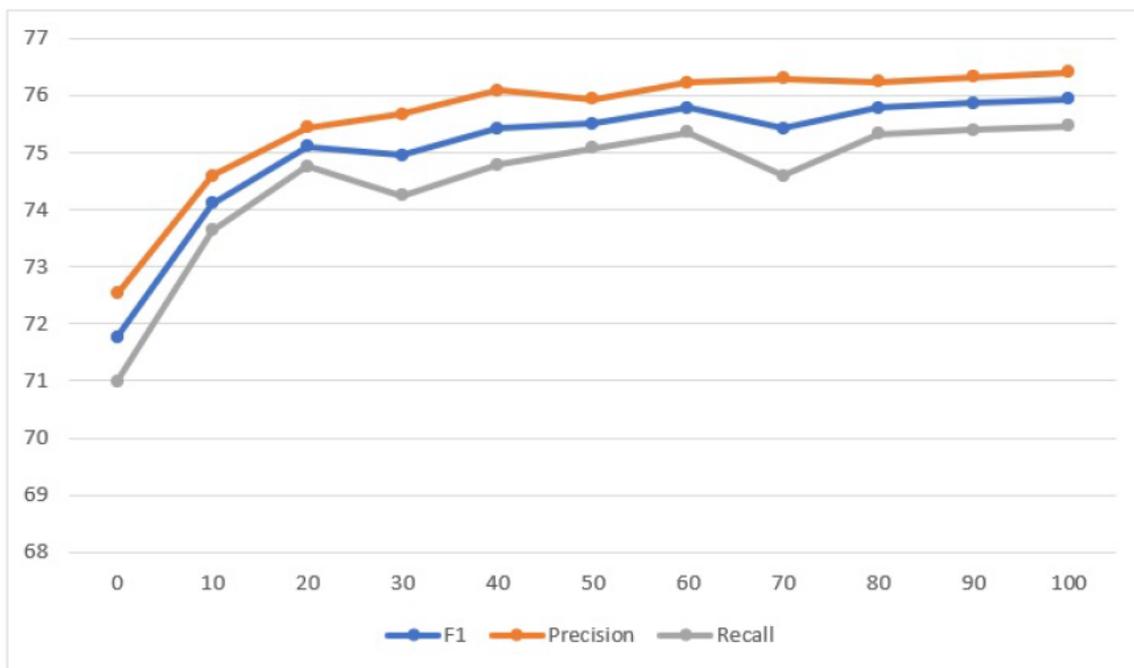

Figure 5: *The trend of F1, Precision, and Recall metrics as the percentage of English data increasess*

**Discussion**

We evaluated our proposed model on the Persian part of the CoNNL-2009 dataset, focusing on cross-lingual semantic role labeling. Since no previous research has been conducted on cross-lingual semantic role labeling in Persian, we compared the proposed model with the model by Nikseirat et al., which achieved the best results in Persian so far. In the proposed model, instead of using the entire training dataset, we only utilized 10% of it for cross-lingual semantic role labeling in both experiments. Following the conventions in English research, we followed a random split for training, validation, and testing, while in Persian, we used a configuration of 10% for testing, 10% for validation, and 80% for training data. The F1 measure in the monolingual mode was 71.76, and in the cross-lingual mode with 10% English data, it was 74.11. Furthermore, if we use all available English data, the F1 measure reaches 75.94. Therefore, this model managed to improve the Persian results by 4.18% compared to the monolingual mode by utilizing English data. By conducting similar experiments using the proposed model by Nikseirat et al., which is the best model for semantic role labeling in Persian, we obtained an F1 measure of 69.71. Thus, our proposed model achieved a 2.05% improvement in the monolingual mode and a 6.23% improvement in the cross-lingual mode compared to this model. However, since our model performs semantic role labeling in all four stages, unlike the assumption made in Nikseirat et al.'s research where only the last two stages are performed, the actual difference between the two models is significantly higher, reaching 6.23 We tried to improve this model by:

- Instead of using language-specific part-of-speech tags, we used universal part-of-speech tags, but we did not see much improvement. Our assumption was that pre-trained models like BERT are capable of extracting features related to part-ofspeech tags.
- Instead of xlm-roberta, we used bertmultilingual- base, which resulted in approximately 10% lower performance.

- Adding more layers to BiLSTM or changing the proposed architecture for the two global encoders was also tested, but it did not help improve the model because the model had reached its maximum learning capacity with the existing number of layers.
- We attempted to use transformers instead of BiLSTM. However, the transformer architecture is only suitable for sequence-to-sequence learning tasks and not for sequence labeling, which is used in the proposed model. Therefore, we were unable to use transformers without modifying the proposed architecture.

This model is the first cross-lingual semantic role labeling model used for the Persian language. However, this model can be used for normal semantic role labeling with all data. The advantage of this model over other proposed models in Persian is that it allows for a unified model for all available languages. In other words, the model is trained in a multitask manner on both Persian and English, and since the parameters for the two main parts of the model, i.e., the universal sentence encoder and the universal predicate-argument encoder, are shared across different languages, the number of learnable parameters is halved compared to training two separate models for two languages.

**Future Work**

The main priority of this model is to simultaneously train on data from multiple languages and prioritize improving the results for a language with limited resources. Therefore, in future work, it can be explored whether the model can be modified to solely focus on improving the results for a language with rich resources, while transferring the performance of the model on data from that language to a secondary priority. One solution we propose for this purpose is to weight the error function for different languages. Additionally, in recent years, new activation functions such as Mish [42] and SERF [43] have been introduced to improve the performance or speed of various tasks. In future work, the effect of replacing Swish with these activation functions or using a combination of them can be investigated. Furthermore, considering that five out of six languages in the CoNNL-2009 dataset are similar to each other, we suggest utilizing data resources for languages like Arabic alongside Persian and English in future work.